\documentclass{article}
\usepackage{spconf,amsmath,graphicx}
\usepackage{xcolor}
\usepackage{adjustbox}
\usepackage{pgf}
\usepackage[pagebackref=true,breaklinks=true,colorlinks,bookmarks=false]{hyperref}

\def\*#1{\mathbf{#1}}
\newcommand\best[1]{\underline{\bf#1}}

\newcommand\task[1]{_{#1}}
\newcommand\ntask{T}
\newcommand\ls{\alpha}
\newcommand\lsp{\ls^{+}}
\newcommand\lsm{\ls^{-}}
\newcommand\smooth[1]{\tilde{#1}}

\newcommand\gt[1]{\textcolor{black}{#1}}

\title{Fighting noise and imbalance in Action Unit detection problems.}
\name{Gauthier Tallec$^{1}$, Arnaud Dapogny$^{2}$, and K\'evin Bailly$^{1,2}$}
\address{$^1$ Sorbonne Universit\'e, 4 Place Jussieu, 75005 Paris, France \\
         $^2$ Datakalab, 114 Boulevard Malesherbes, 75017 Paris, France\\}

\begin{document}
%
\maketitle
\begin{abstract}
      Action Unit (AU) detection aims at automatically caracterizing facial expressions with the muscular activations they involve. Its main interest is to provide a low-level face representation that can be used to assist higher level affective computing tasks learning. Yet, it is a challenging task. Indeed, the available databases display limited face variability and are imbalanced toward neutral expressions. Furthermore, as AU involve subtle face movements they are difficult to annotate so that some of the few provided datapoints may be miss-labeled. In this work, we aim at exploiting label smoothing ability to mitigate noisy examples impact by reducing confidence \cite{lukasik2020does}. However, applying label smoothing as it is may aggravate imbalance-based pre-existing under-confidence issue and degrade performance. To circumvent this issue, we propose Robin Hood Label Smoothing (RHLS). RHLS principle is to restrain label smoothing confidence reduction to the majority class. In that extent, it alleviates both the imbalance-based over-confidence issue and the negative impact of noisy majority class examples. From an experimental standpoint, we show that RHLS provides a free performance improvement in AU detection. In particular, by applying it on top of a modern multi-task baseline we get promising results on BP4D and outperform state-of-the-art methods on DISFA.

\begin{figure}[t!]
    \centering
    \includegraphics[width=8.6cm]{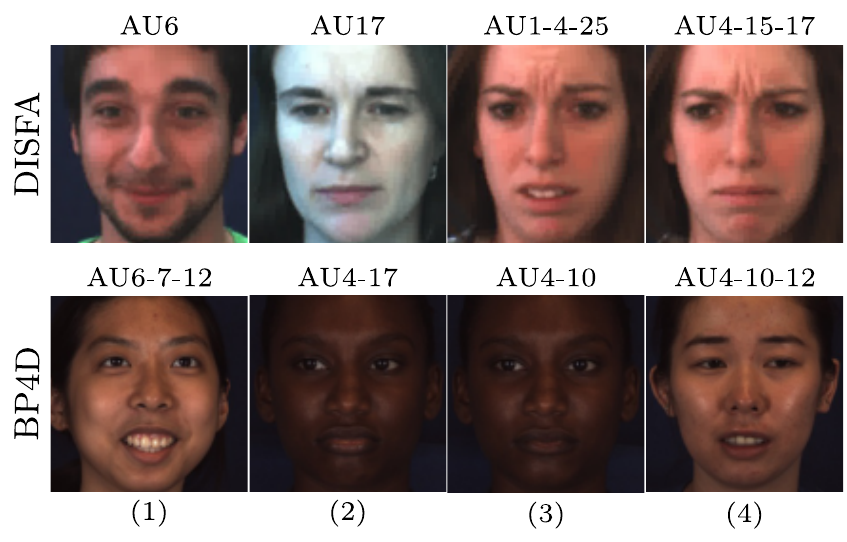}
    \caption{Noisy examples from DISFA and BP4D. For example, the first face in DISFA is annotated with neither smile (AU12) or eyebrow raise (AU1-2). Fitting on those examples may prevent the network from properly understanding which zones are involved in each AU and degrade performance.}
    \label{fig:au_noise}
\end{figure}
\end{abstract}
\begin{keywords}
Computer Vision, Affective Computing, Action Unit Detection.
\end{keywords}

\section{Introduction}

Facial expressions convey information galore about how humans feel. Consequently,  efficient computational face representation could unlock better automatic comprehension of human behaviours and in turn improve human-machine interaction. For that purpose, the Facial Action Coding System (FACS) provides an anatomic representation that decompose faces using muscular activations called Action Units (AU).

From a machine learning point of view, AU detection can be formulated as a multi-task problem where each task consists in the detection of a single AU. In practice, its performance are hindered by data scarcity. Indeed, the AU labeling process consists in frame-by-frame video annotations of subtle facial activations and is therefore hardly scalable. As a result existing AU datasets display low face variability with only few  positively annotated examples (because AU are short events). Finally, as shown in Figure \ref{fig:au_noise}, AU are often so subtle that even expertly trained annotators may miss-annotate edgy examples \cite{zhang2014bp4d}, resulting in label noise. Altogether, training in such setting is prone to overfitting and model over-confidence toward predicting neutral expressions.  

To tackle low face variability, the main line of research make use of prior geometric information (typically facial keypoints) to help the learning process by either guiding the network attention  \cite{li2018eac, shao2018deep, bonnard2022privileged, shao2020jaa} or normalizing face geometry \cite{niu2019local}. In the same vein , several works \cite{tallec2022multi, corneanu2018deep, song2021hybrid} attempted to incorporate prior AU dependencies to better structure predictions. For imbalance problems, the widely adopted approach is loss frequency reweighting \cite{shao2019facial, jacob2021facial, shao2018deep, shao2020jaa}. 

Interestingly, very few methods address the label noise problem. The work in \cite{song2021uncertain} is among the only that takes AU uncertainty into account by using the method in \cite{kendall2018multi} to learn to adapt the contribution of each AU to the total loss. 

This work lies in the continuity of \cite{song2021uncertain} since it focuses on AU noise modelling. Yet, we stand aside from it arguing that uncertainty learning intuitively requires large amount of data \cite{sukhbaatar2014training} and may therefore not be fully efficient in AU detection low data regime. 

Instead, we aim at taking advantage from the recent success of label smoothing \cite{szegedy2016rethinking} at mitigating noise \cite{lukasik2020does} by reducing over-confidence. However, vanilla label smoothing reduces over-confidence uniformly in all classes. Therefore, applying it in imbalanced situations may worsen the pre-existing under-confidence in the minority class. For that purpose, we propose Robin Hood Label Smoothing (RHLS) that takes its name from the fact that, by smoothing only the majority class, it introduces a probability to take examples from the rich class to give them to the poor. By doing so, it reduces both the imbalance-based over-confidence issue and the negative impact of noisy majority class examples. To summarize our contributions are as follows :
\begin{itemize}
    \item We introduce RHLS that adapts label smoothing to imbalanced situations by restraining over-confidence reduction to the majority class. Consequently it mitigates both imbalance over-confidence issue and the negative impact of majority class noisy examples.
    \item Experimentally, we show that AU detection performance benefits from the use of RHLS without any additional computational overhead. More precisely we demonstrate that applying RHLS on a modern multi-task baseline is competitive on BP4D and significantly outperforms state-of-the-art results on DISFA. 
\end{itemize}

\section{Methodology}

\begin{figure}
    \centering
    \includegraphics[width=8.6cm]{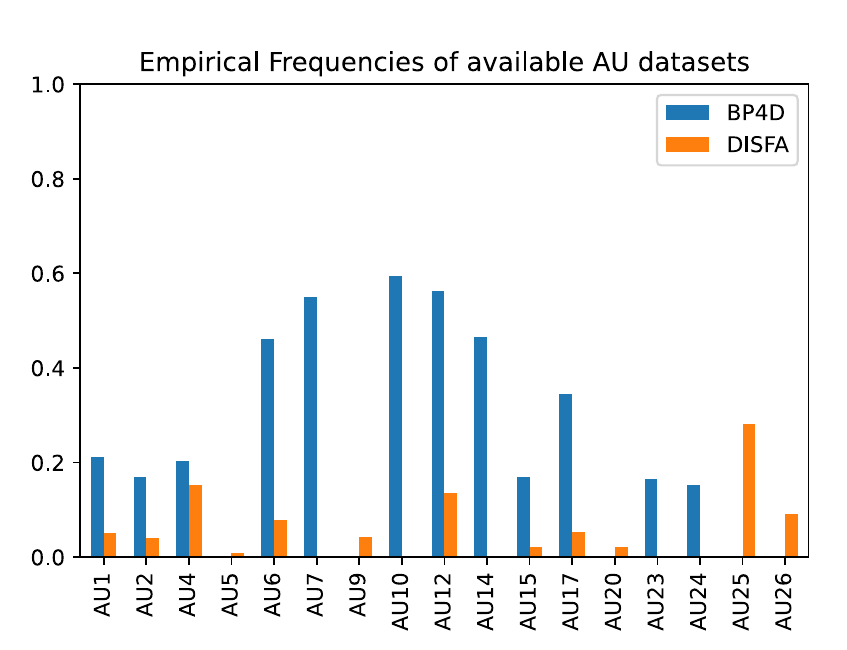}
    \caption{Action Unit frequencies for BP4D and DISFA. BP4D is slightly imbalanced toward negativeness compared to DISFA where most AU are represented in less than 1/10 frame. Minimizing cross-entropy in such imbalanced situations tend to push the network toward over-confidence in the majority class i.e toward ignoring the minority class and predicting the majority class with high probability.}
\label{fig:empirical_frequencies}
\end{figure}

In this work, we use a multi-task binary classification dataset composed of couples $(\*x, \*y)$ with $\*x\in\mathbf{R}^{H\times W \times3}$ a face image and $\*y \in \{0, 1\}^{\ntask}$ the labels for each of the $\ntask$ target AU.

\subsection{Vanilla Label Smoothing}
AU detection involves subtle changes in skin texture that are difficult to detect, even for expertly trained annotators. As a consequence, the main available annotated datasets display noise. Figure \ref{fig:au_noise} highlights the existence of this noise by showing several arguably wrong annotated examples. Prior work \cite{lukasik2020does} showed that label smoothing could help mitigate the influence of annotation noise by reducing model confidence \cite{szegedy2016rethinking} and consequently preventing it from over-fitting on noisy examples. For that purpose, label smoothing introduces uniform noise into the ground truth labels with probability $\ls$. From a concrete point of view, label smoothing with coefficient $\ls$ modifies ground truth  label of task $i$ as follows: 

\begin{equation}
\smooth{y}\task{i} = (1 - \ls) y\task{i} + \frac{\ls}{2}.
\end{equation}

However, we experimentally show that label smoothing degrades AU detection performance (see section \ref{subsec:ablation_study}). We hypothesize that such performance drop is due to AU datasets imbalance. Indeed, as shown by figure \ref{fig:empirical_frequencies}, in nowadays most popular AU datasets, several AU display low empirical frequencies. In particular, such imbalance has been shown \cite{lin2017focal} to push model toward under-confident predictions for the minority class. Therefore, by reducing confidence of both positive and negative examples, label smoothing may worsen the pre-existing confidence problem on the minority class and in turn explain the observed performance gap. 

\subsection{Robin Hood Label Smoothing (RHLS)}

In order to address that problem we extend label smoothing to Robin Hood Label Smoothing (RHLS). RHLS takes its name from the fact that, by smoothing only the majority class, it introduces a probability to steal examples from the majority (the rich) class to give it to the minority class (the poor). Formally, it first introduces $\lsp\task{i}$ and $\lsm\task{i}$ that respectively denote the uniform noise probability for positive and negative values for task $i$ so that :

\begin{align}
\smooth{y}\task{i} = y\task{i}(1 - \frac{\lsp\task{i}}{2}) + (1 
 - y\task{i})\frac{\lsm\task{i}}{2}.
\end{align}

Then it parametrizes $\lsp\task{i}$ and $\lsm\task{i}$ with respect to task $i$ empirical frequencies $f_i$ so that only the majority class is smoothed :

\begin{equation}
\lsm\task{i} = \beta \max(0, \frac{1 - 2 f\task{i}}{1 - f\task{i}}), \lsp\task{i} = \beta \max(0, \frac{2 f\task{i} - 1}{f\task{i}}),
\end{equation}

where $\beta \in [0, 1]$ quantifies the amount of noise introduced in the majority class from $0$ (No noise is applied) to $1$ (noise is applied so that the resulting dataset is balanced).

Through the introduction of noise in the majority class using $\beta > 0$, RHLS encourage less-confident prediction for negative examples. It consequently reduces the negative impact of noisy negative examples, alleviates the imbalance-based overconfidence problem an may improve performance w.r.t vanilla label smoothing.

In what follows, we validate the RHLS superiority over vanilla label smoothing in imbalanced situations and discuss the significant improvements it provides in AU detection.


\section{Experiments}

\begin{figure}
    \centering
    \includegraphics[width=8.6cm]{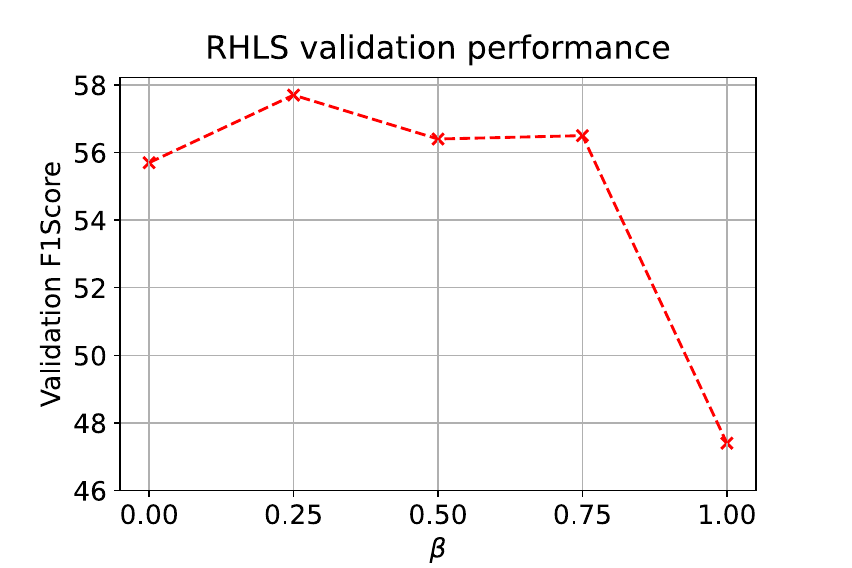}
        \caption{RHLS $\beta$ validation \label{tab:rhls_valid} on DISFA.}
    \label{fig:rhls_valid}
\end{figure}

 	


In this section, we first introduce Action Unit Detection datasets
(Section \ref{subsec:datasets}) along with details about our architecture and its optimization (Section \ref{subsec:implem}). Then, in Section \ref{subsec:ablation_study} we both validate our method hyperparameters and perform a comparative analysis between RHLS and vanilla methods for fighting noise and imbalance. Finally In Section \ref{subsec:sota_comparaison},
we compare RHLS with state-of-the-art approaches.

\subsection{Datasets}
\label{subsec:datasets}

\textbf{DISFA} \cite{mavadati2013disfa} is a dataset for facial action unit detection. It consists of 27 single subject videos for a total of $\approx 100k$ face images. Each image is annotated with 12 AU intensity scores that range from 0 to 5. In detection, intensity scores higher than 2 are considered positive \cite{zhao2016deep}. As far as evaluation is concerned, the 27 videos are split into 3 folds of 9 \cite{shao2018deep} and performance is measured by averaging the 3 mean F1-Scores obtained from training on 2 folds and evaluating on the last. For stability concerns \cite{shao2018deep, shao2019facial}, we run such evaluation protocol 5 times and report mean performance. For validation, we follow the protocol in \cite{shao2018deep} i.e we perform 6 fold cross-validation on each of the 3 two-folds training set and compute the validation scores by averaging F1-Score on those 18 runs.

\textbf{BP4D} \cite{zhang2014bp4d} dataset is composed of approximately $140k$ face images in which $41$ people ($23$ females, $18$ males) with different ethnicities are represented. Each image is annotated with the presence of $12$ AU. Similarly to DISFA, performance evaluation consists in measuring  F1-Scores on all $12$ AUs using a subject exclusive 3-fold cross evaluation with the same fold distribution as in \cite{shao2018deep}.

\subsection{Implementation Details}
\label{subsec:implem}

 For all our experiments, we inspire from the M architectures in \cite{touvron2021going}. However, as no face-based pretrained ViT is publicly available, we replace this part by a resnet50 pretrained on VGGFACE \cite{cao2018vggface2}. To fit with the ViT encoder outputs in \cite{touvron2021going} that is $P \times d$ where $P=196$ and $d=768$, we replace the last convolutional block by a conv1D of size $d=768$ on top of which a layer of self-attention is built. For the decoder part we use as many class tokens in the cross-attention as there are AU in the dataset ($T=8$ for DISFA and $T=12$ for BP4D) and we feed each of the resulting $T$ representations to an AU specific dense layer with sigmoidal activation.

For optimization we use AdamW \cite{adamw} with exponential decay $\beta = 0.75 $ for $2$ epochs. In the convolutional part we use initial learning rate $\lambda_c = 5e-5$ for BP4D and $\lambda_c = 1e-5$ for DISFA. In the transformer part we scale the initial learning rate w.r.t number of queries $q$ ($P$ in self attention and $T$ in cross-attention), model size $d$ and batchsize $B = 32$, so that $\lambda_t = \lambda_t^{(0)}\frac{Bq}{\sqrt{d}}$ and $\lambda_t^{(0)} = 4e-8$.

\subsection{Ablation Study}
\label{subsec:ablation_study}

In this section we validate smoothing intensity $\beta$ and compare RHLS to  vanilla existing methods for noise mitigation (label smoothing \cite{szegedy2016rethinking}) and imbalance (frequency weighted cross-entropy \cite{shao2018deep, shao2019facial, jacob2021facial}) on DISFA.

Figure \ref{tab:rhls_valid} shows the evolution of the validation score with respect to $\beta$. For low values of $\beta$, RHLS significantly boosts the model predictive performance by reducing overconfidence in the majority class and consequently lowering the negative influence of both imbalance and noisy negative examples. However, passed a certain threshold for $\beta$, RHLS introduces too much false positive in training which hurts the learning process and results in performance drops. Therefore, we select $\beta=0.25$ for evaluation. 

\begin{figure}
    \includegraphics[width=8.6cm]{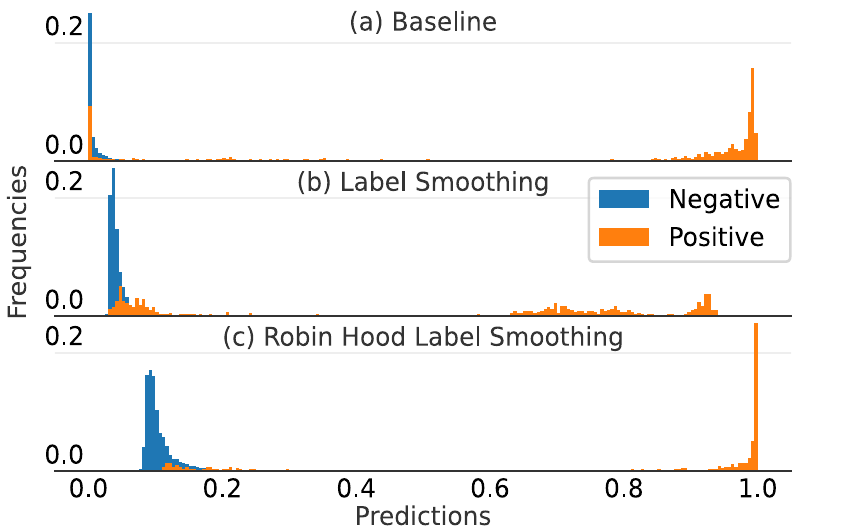}
        \caption{Histogram of predictions on positive and negative DISFA AU2 labels for different smoothing methods.\label{tab:fig:histo_pred}}
    \label{fig:histo_pred}
\end{figure}

Second, Table \ref{tab:ablation_study} compares the performance of the proposed RHLS with other existing label smoothing methods. First, it is noticeable that frequency weighted BCE hurts performance. This may be caused by the difference of scales across AU frequencies (eg : $f_{\text{AU1}} \sim 1 e - 2$, while $f_{\text{AU25}} \sim 1 e - 1$) so that weighting AU loss contribution using $w_i = \frac{1/f_i}{\sum_{j=1}^{T} \frac{1}{f_j}}$ may encourage the learning of extremely low frequency AU at the expense of all the others.

\gt{Finally, figure \ref{fig:histo_pred} show the histogram of predictions for different smoothing method on DISFA. We observe that baseline results display majority class overconfidence as many positive examples are predicted negative with probability $0$. Label smoothing mitigates that problem but worsens the imbalance-based minority class low confidence problem which in turn reduce overall performance (see Table \ref{tab:ablation_study}). By smoothing only the majority class, RHLS mitigates majority class confidence without any influence on the minority class and consequently obtains the performance boost reported in Table \ref{tab:ablation_study}.}

\subsection{Comparaison with state-of-the-art methods}
\label{subsec:sota_comparaison}

In this section we compare RHLS with state-of-the-art AU detection methods.

Table \ref{tab:sota_summary} provides RHLS results on BP4D. Interestingly, applying RHLS over a modern multi-task baseline is competitive with several recent methods \cite{shao2019facial, shao2020jaa} including other uncertainty modelling strategies \cite{song2021uncertain}. However, it gets outperformed by the most recent ones that either involve more complex landmark guided transformer architecture \cite{jacob2021facial} or refined AU dependency modelling \cite{tallec2022multi, song2021hybrid}. Nonetheless, the increment RHLS provides to the baseline shows that it is a simple yet efficient way to improve performance without any additional computational overhead. In that extent, attempting to plug it on top of more complex methods could be a promising track toward better overall AU detection performance. 

\begin{table}[t!]
    \centering
    \resizebox{0.8\linewidth}{!}{
\begin{tabular}{|c|c|c|c|c|c|c|}
\hline
Method & Mean F1Score \\
\hline
Baseline & 63.0 $\pm$ 1.9 \\
Label Smoothing($\alpha = 0.1$) & 62.0 $\pm$ 1.8 \\
Frequency Weighted BCE & 61.7 $\pm$ 2.1 \\
Robin Hood Label Smoothing & 65.8 $\pm$ 1.4 \\
\hline
\end{tabular}}
    \caption{Comparaison between RHLS and prior label smoothing methods for noise and imbalance mitigation on DISFA dataset. \label{tab:ablation_study}}
\end{table}

On DISFA, Table \ref{tab:sota_summary} shows that the free increment RHLS allows the baseline architecture to surpass state-of-the-art performance. To explain those excellent results, it is worth noticing that most state-of-the-art methods \cite{shao2018deep, shao2019facial, tallec2022multi, jacob2021facial} use frequency weighted loss. Therefore the superiority of RHLS against the frequency weighted loss on DISFA (see Table \ref{tab:ablation_study}) may justify the significant increment that we observe.

\begin{table}[t!]
    \centering
	
\begin{tabular}{|c|c|c|}
\hline
\footnotesize {\bf{Mean F1 Score}} & \bf{BP4D} & \bf{DISFA} \\
\hline
\footnotesize JAANet \cite{shao2018deep} & 60.0 & 56.0 \\
\footnotesize ARL \cite{shao2019facial} & 61.1 & 58.7 \\
\footnotesize JÂANET \cite{shao2020jaa} & 62.4 & 63.5\\
\footnotesize UGN-B \cite{song2021uncertain} & 63.3 & 60.0 \\
\footnotesize FAUwT \cite{jacob2021facial} & 64.2 & 61.5\\
\footnotesize MONET  \cite{tallec2022multi} & \best{64.5} & 63.9\\
\hline
\footnotesize Baseline & 61.9 & 63.1 \\
\footnotesize RHLS & 63.0 & \best{65.8} \\
\hline 
\end{tabular}
    \vspace{0.1cm}
    \caption{Comparaison of RHLS with state-of-the-art deep learning based AU detection methods \label{tab:disfa_results}}
    \label{tab:sota_summary}
\end{table}
Beyond that, it is also worth noticing that RHLS simple and free label modification pushes a simple baseline above more complex methods with spatial prior guidance \cite{shao2020jaa} or explicit AU dependency modelling \cite{tallec2022multi}. On the one hand, it highlights that imbalance reduction and noise modelling are 
 as critical as input feature extraction or dependency modelling in AU detection performance. On the other hand, it offers potential improvement perspective as integrating RHLS with more complex methods could further improve performance.

\section{Conclusion}

\gt{In this work, we investigated the impact of label smoothing to fight against AU datasets imbalance and noise problems. In particular, we showed that vanilla label smoothing is ill-adapted to imbalanced situations as it may worsen pre-existing under-confidence problems and degrade performance. To alleviate this issue, we proposed Robin Hood Label Smoothing that constrains label smoothing to the majority class by introducing a probability to steal examples from the rich (the majority class) to give it to the poor (the minority class). In that extent RHLS reduces both imbalance issues and majority class noise negative impact}

\gt{Experimentally, we showed that applying RHLS on top of a multi-task baseline provides competitive performance on BP4D and significantly outperforms state-of-the-art on DISFA. In particular, on DISFA, the excellent results obtained indicates that RHLS is a better option that the frequency weighted loss. In future works, we will inspire from the successes of AU dependency modelling methods \cite{song2021hybrid, tallec2022multi} and try structuring label smoothing noise to prevent smoothed labels from displaying unrealistic dependencies that may hurt the training process.}  



\label{sec:intro}

\section{Acknowledgements}
This work was granted access to the HPC resources of IDRIS under the allocation 2021-AD011013183 made by GENCI.



\vfill\pagebreak

\label{sec:refs}

\bibliographystyle{IEEEbib}
\bibliography{refs}

\end{document}